\documentclass[sigconf, nonacm]{acmart}

\usepackage[a-1b]{pdfx}
\usepackage{tabularx}
\usepackage{amsmath}
\usepackage[noend]{algpseudocode}
\usepackage{algorithm}
\usepackage{amsthm}
\usepackage{amsfonts}
\usepackage{graphicx}
\usepackage{subfig}
\usepackage{url}
\usepackage{caption}
\usepackage{balance}
\usepackage{hyperref}
\usepackage{xcolor}
\usepackage{xspace}
\usepackage{stfloats}
\usepackage{float}
\usepackage{bm}
\usepackage{enumitem}
\usepackage{comment}
\newcommand{\eat}[1]{}

\newcommand\vldbdoi{10.14778/3523210.3523219}
\newcommand\vldbpages{1417-1425}
\newcommand\vldbvolume{15}
\newcommand\vldbissue{7}
\newcommand\vldbyear{2022}
\newcommand\vldbauthors{\authors}
\newcommand\vldbtitle{\shorttitle} 
\newcommand\vldbavailabilityurl{URL_TO_YOUR_ARTIFACTS}
\newcommand\vldbpagestyle{empty} 

\begin{document}
\title{ OnlineSTL: Scaling Time Series Decomposition by 100x }

\author{Abhinav Mishra }
\affiliation{
    \institution{Splunk}
    \city{San Francisco}
    \country{US}
    }
\email{amishra@splunk.com}

\author{Ram Sriharsha }
\affiliation{
    \institution{Pinecone}
    \city{San Francisco}
    \country{US}
    }
\email{ram@pinecone.io}

\author{Sichen Zhong }
\affiliation{
    \institution{Splunk}
    \city{San Francisco}
    \country{US}
    }
\email{szhong@splunk.com}

\begin{abstract}
Decomposing a complex time series into trend, seasonality, and remainder components is an important primitive that facilitates time series anomaly detection, change point detection, and forecasting. Although numerous batch algorithms are known for time series decomposition, none operate well in an online scalable setting where high throughput and real-time response are paramount. In this paper, we propose OnlineSTL, a novel online algorithm for time series decomposition which is highly scalable and is deployed for real-time metrics monitoring on high-resolution, high-ingest rate data.
Experiments on different synthetic and real world time series datasets demonstrate that OnlineSTL achieves orders of magnitude speedups (100x) for large seasonalities while maintaining quality of decomposition. 

\end{abstract}

\maketitle

\pagestyle{\vldbpagestyle}
\begingroup\small\noindent\raggedright\textbf{PVLDB Reference Format:}\\
\vldbauthors. \vldbtitle. PVLDB, \vldbvolume(\vldbissue): \vldbpages, \vldbyear.\\
\href{https://doi.org/\vldbdoi}{doi:\vldbdoi}
\endgroup
\begingroup
\renewcommand\thefootnote{}\footnote{\noindent
This work is licensed under the Creative Commons BY-NC-ND 4.0 International License. Visit \url{https://creativecommons.org/licenses/by-nc-nd/4.0/} to view a copy of this license. For any use beyond those covered by this license, obtain permission by emailing \href{mailto:info@vldb.org}{info@vldb.org}. Copyright is held by the owner/author(s). Publication rights licensed to the VLDB Endowment. \\
\raggedright Proceedings of the VLDB Endowment, Vol. \vldbvolume, No. \vldbissue\ %
ISSN 2150-8097. \\
\href{https://doi.org/\vldbdoi}{doi:\vldbdoi} \\
}\addtocounter{footnote}{-1}\endgroup


\keywords{Time series analysis, Online algorithm, High-performance systems}

\section{Introduction}

Distributed systems and microservice-based architectures have evolved significantly over the last decade. They produce numerous telemetry data as well as data coming from containers, servers, and devices. Today, cloud operators which provide services for monitoring these {\it metrics} include Amazon CloudWatch \cite{cloudwatch},  Datadog \cite{ddog}, Google StackDriver \cite{stackdriver}, Microsoft Azure Monitor \cite{azure}, and Splunk \cite{splunk}. Metrics such as network latency, disk utilization, CPU usage, memory usage, and incoming request rate are collected from these entities for monitoring. In fact, there are many databases aimed at storing time series metrics such as Google Monarch \cite{monarch}, Graphite \cite{graphite}, OpenTSDB \cite{opentsb}, and Prometheus \cite{prometheus}.

Monitoring metrics is a fundamental task of DevOps \cite{asap, devops1, devops2}. This
includes continuous monitoring of system health on a dashboard, detecting anomalies for
any abnormal behavior, alerting, and querying to obtain aggregated reports. For example, if the metric "disk utilization" is high, then this may indicate the server will reject future requests. At the same time, there has been an explosion in the cardinality of metrics being monitored in stream over the last few years \cite{uber, monarch}.
Reducing the dimensionality of metrics automatically to deal with this sort of explosion in cardinality is an active area of research \cite{sieve}.
Moreover, metrics are being ingested at ever-increasing rates and granularities; Facebook \cite{fb}, LinkedIn\cite{linkedin}, and Twitter\cite{twitter}, for example, generate over 12M events per second.

For DevOps at Splunk, monitoring cost grows linearly with number of events per second across all metrics. For example, throughput of 100 
records/sec may require 120,000 CPUs for 12M events/sec.
Now, if throughput is increased to 10,000 records/sec, then the respective
CPU requirement can be reduced to 1,200 CPUs, drastically lowering cost. In the standard setting,
DevOps mitigates these costs by either storing data at lower resolutions (for example, hourly instead of minutely) or running algorithms less frequently, which result in non-real-time monitoring as well as computational overheads such as maintaining aggregates to keep track of incoming data. 
Our approach is to design scalable algorithms with high throughput. Specifically in this work, we introduce scalable methods based on fast incremental updates for time series decomposition which are 
fundamental for efficient metrics monitoring.

\begin{figure}[tb]
\centering
\includegraphics[scale=0.44]{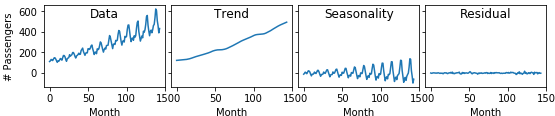}
\caption{Decomposition of the Airline passenger data \cite{boxjen76} from 1949 to 1960. Notice that trend is a 
smooth long pattern, where as seasonality is a repeated pattern
in the data.}
\label{Fig: stl_decomp}
\end{figure}

Trend and seasonal patterns exist in almost every one of these metrics. Certain parts of the day see different patterns. For example, higher network latency is not unexpected during peak hours. If the observed network latency is low during peak hours, it might imply that many users are not able to use the service, and the system should raise an alarm.
Almost all cloud operators account for seasonality when evaluating time series \cite{amazon-anomaly, azure-anomaly, datadog-anomaly}. An example of a time series decomposition is shown
in Figure \ref{Fig: stl_decomp}. 

In order to correctly account for seasonality, one typically performs time series decomposition to subtract out the seasonal and trend components. The residual is then used to monitor for outliers. For this sort of approach to work at scale, the decomposition needs to handle high cardinalities and ingestion rates. However, this poses significant challenges to all existing time series decomposition algorithms including classical STL \cite{stl}, SEATS \cite{seats}, and TBATS \cite{tbats}, as well as more recent algorithms such as STR \cite{str} and RobustSTL \cite{robuststl}. Another issue is that existing approaches are batch algorithms which work on the entire dataset and do not have mechanisms for fast updating given new incoming data points, i.e., these algorithms are not online and cannot function on unbounded data streams.

Ideally, we desire a high-performance algorithm on a distributed platform which supports unbounded streams such as Apache Flink \cite{flink}. To the best of our knowledge, there do not exist time series decomposition algorithms which work on streams, are online, and are also deployed on such distributed systems. In this work, we present OnlineSTL, a scalable online algorithm which emphasizes high throughput and is deployed for real-time DevOps metrics monitoring. We propose an algorithm with the following contributions:

\begin{itemize}[leftmargin=*]
    \item first online time series decomposition algorithm. We achieve this using fast incremental updates. This results in significantly improved throughput for time series exhibiting large seasonalities, thus lowering the overall data center cost.
    \item ability to ingest high-resolution data and handle long seasonalities (over 10,000). We can ingest data arriving at a resolution of seconds/minutes.
    \item successfully deployed on Apache Flink \cite{flink} for over a year and applied to real-time metrics monitoring use cases handling hundreds of thousands of entities. 
    \item maintains a competitive decomposition quality versus other state of the art algorithms.
\end{itemize}

As we can see from Table \ref{table:algorithm_comparison}, existing time series decomposition techniques force us into a severe tradeoff: either increase resources and hence cost significantly more to keep up with volume and cardinality, or lower the resolution of the time series (or equivalently increase the staleness of the trained models) in order to create a sustainable throughput. Furthermore, OnlineSTL has a dramatic increase in throughput on high ingest, high seasonality datasets compared to the second best algorithm, which happens to be "classical" STL \cite{stl}. Apart from low throughout of these existing batch algorithms, an additional challenge is the length of periodicity in the data. All of these algorithms are too slow when resolution of data is aggregated by minutes or seconds. Many existing techniques were designed to be used in lower resolution data, like those seen in econometrics and finance. This explains why popular techniques like SEATS \cite{seats} and TBATS \cite{tbats} only scale up to resolutions of daily or weekly, and are not designed to handle long seasonalities which are common in DevOps use cases.

\begin{table}[t]
\centering
\caption{OnlineSTL is the only algorithm that is both online
	and can handle long seasonality periods. Throughput is the number of data points processed per sec by the algorithm on a single core. Data arrives at a minutely resolution for a weekly seasonality of 1440.
	}
\label{table:algorithm_comparison}
{\fontsize{8}{8}\selectfont
\begin{tabular}[l]{ ccccc } 
\toprule
 Algorithm   & Online & Long  & Multiple & Throughput \\
 & &  Seas. &  Seas. & per sec \\
\midrule
	STL \cite{stl}/MSTL \cite{mstl}  & N & N & Y & O(100) \\

	TBATS \cite{tbats}& N & N & Y & O(1) \\
	STR \cite{str} & N & N & Y & O(1) \\
	SSA \cite{ssa} & N & N & Y & O(1) \\
	RobustSTL \cite{robuststl} & N & N & N & O(1) \\
	Fast-RobustSTL \cite{frobustSTL} & N & N & Y & O(1) \\
	\bf{OnlineSTL} & \bf{Y} & \bf{Y} & Y & \bf{O(10,000)} \\
\bottomrule
\end{tabular}
}
\end{table}
\raggedbottom

Internally, we have deployed OnlineSTL for over a year now, and the algorithm is implemented on top of Apache Flink \cite{flink} as a stateful keyed map function. Apache Flink is a distributed streaming data-flow engine. It remains a natural choice for systems needing low latency and high throughput. In Flink, data is streamed continuously, queries are never-ending, and data can arrive late or even out-of-order.

\section{Related Work}
Time series decomposition has been extensively studied in time series analysis and econometrics. One of the earliest methods was discovered in 1884 by Poynting \cite{pointing}, for removing the
trend and seasonality on pricing data. Variants and improvements to Arima include the classical methods proposed by Macaulay \cite{Macaulay} and the more recently developed X-13 ARIMA-SEATS \cite{finley}.

Seasonal-Trend decomposition via Loess (STL) was developed by Cleveland et al \cite{stl} and has been the most popular technique. This is due to the the algorithm supporting non-linear change in trend and seasonality. MSTL \cite{mstl, bandara2021mstl}, the extension of STL to support multiple seasonalities, was later proposed. 
In STL, trend and seasonal components are computed alternately. 
Loess smoothing is then applied to extract trend, and is also used to obtain changing seasonal components. STL is less flexible in many ways, such as its inability to handle seasonality shifts, large periodicity, and high noise. 
Other model-based approaches such as SEATS \cite{seats} and TBATS \cite{tbats} provide confidence and prediction intervals as well. Extensions to these techniques make them robust and popular in many different fields.

Seasonal-Trend decomposition using regression (STR) \cite{str}  jointly computes trend and seasonality components by learning a two-dimensional structure. 
It also allows shifts in seasonality as well as resilience to outliers. Robustness is achieved using $l_1$-norm regularization. However, the algorithm does not scale because learning two-dimensional structure is computationally expensive. Furthermore, there are many approaches based on matrix decomposition, such as SSA \cite{ssa} (based on SVD). SSA forms a matrix by folding a time series, followed by applying PCA. 
More recently, RobustSTL \cite{robuststl} was introduced. RobustSTL can handle abrupt changes in trend and performs well in the presence of outliers. 
Initially, trend is extracted by optimizing a least-absolute deviation regression objective with sparse regularization. A follow-up on RobustSTL allows for multiple seasonalities \cite{kdd2021}. All of these existing algorithms are batch algorithms, meaning that they must be retrained for every new data point, fundamentally limiting scalability.
One could hope to rectify this limitation by retraining periodically, but this subjects an algorithm to failure resulting from concept drift or staleness. 

\section{Preliminaries} \label{prelim}
\subsection{Definitions}
Before we jump into our algorithm and results, we will need some definitions. Let $X = \{X_1, X_2, \cdot \cdot \cdot, X_n\}$ be an unbounded stream of events where $X_j$ represents the value at time $j$. For simplicity, we assume timestamps increase in increments of one. Here, $X_1$ is the earliest value while $X_n$ is the latest value. In this work, we focus on streaming data. Hence, data arrives continuously and is unbounded. In the online setting, each data point arrives sequentially, that is $X_1$ arrives first, then $X_2$, and so on. For any point $X_i$, we want to decompose it into trend, seasonal, and residual components. In this work, we only consider additive decomposition. In other words, $X_i = T_i + \sum_{p=1}^{k}{S_{p, i}} + R_i$, where $T_i$, $S_{p, i}, 1 \le p \le k$ and $R_i$ are the trend, seasonal, and residual components respectively. Therefore, $X$ is decomposed into the unbounded components: trend $T = \{T_1, T_2, \cdot \cdot \cdot, T_n\}$, seasonal components $S_p = \{S_{p, 1}, S_{p, 2}, \cdot \cdot \cdot, S_{p, n}\}$, and residual $R = \{R_1, R_2, \cdot \cdot \cdot, R_n\}$. 

Intuitively, trend captures long-term progressions in the series and is assumed to be locally smooth. Seasonal components reflect repeated periodic patterns, and residual captures all unexplained patterns which are not captured by either trend or seasonal components. 
Throughout this paper, seasonality and period are used interchangeably. We say that $X$ has seasonalities or periods $|S_p| := m_p, 1 \le p \le k$ if the  data contains periods of $m_1, m_2, ..., m_k$. For example, if $X$ is hourly data, then $X$ could have daily seasonality of $m_1 = 24$, and a weekly seasonality of $m_2 = 168$.

\subsection{Batch vs Online Framework}
All existing batch and online decomopositions use a combination of trend filters, seasonality filters, denoising filters \cite{robuststl}, differencing filters, or optimization filters \cite{prophet, robuststl, kdd2021}.
The exact implementation of each approach varies wildly. For example, RobustSTL uses an optimization-based approach to extract trend, while STL uses local linear regression to estimate trend. Facebook's Prophet \cite{prophet} jointly estimates trend and seasonality by solving an optimization problem. Joint estimation typically
involves multiple iterations over the dataset and is  inefficient in an online setting.

 
{\bf Batch Framework:}
Algorithms such as STL, RobustSTL, TBATS can all be seen as a series of filters applied on the whole dataset. Each filter takes the complete output of the previous filter and produces a new vector, which becomes an input to the next filter. Such approaches are usually slow, in that the filters need to act on the entire dataset to output a vector that is linear in the size of the entire dataset. This batch processing becomes the bottleneck in the online setting. In designing an online algorithm, we need to avoid filters that act on large subsets of the stream. In an online setting where the data stream is unbounded, batch filters are considered unacceptable because the algorithm has to restart with every new data point. For example, RobustSTL contains a denoising filter, which is followed by an optimization filter to extract trend, and then is finally followed by a seasonality filter. When a new point arrives, weights output by the denoising filter will change, so we need to recompute the results of the denoising filter on all of  previous points. This attribute is extremely undesirable in the online setting. 

{\bf Online Framework:}
An online filter is required to accept a point and return the result immediately without waiting for future points or recomputing on past points. An example is the moving average trend filter. It is easy to maintain the moving average online since we do not need to recompute over previous points to obtain the new mean. Similar principles apply to any online framework for decomposition. Each filter must accept a single point as input, update some summary statistics quickly, and then immediately pass the result to the next filter. In a data streaming paradigm, an algorithm must use sub-linear (poly-logarithmic) space in the number of instances it has observed. In case of OnlineSTL, we use O(4*max seasonality) space, which is independent of number of data points seen.

\begin{figure*}[t]
\centering
\includegraphics[scale=0.52]{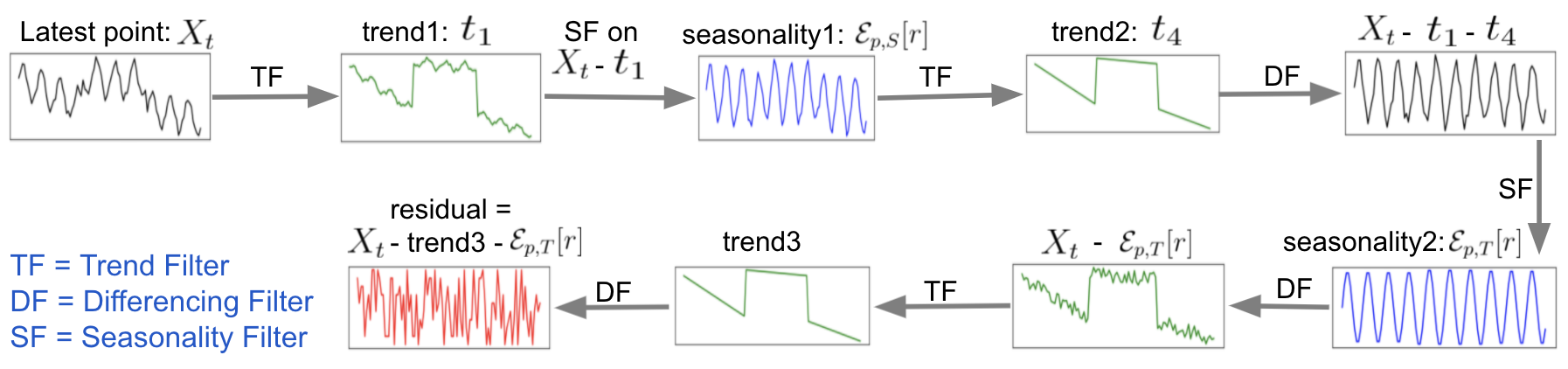}
\caption{An example of OnlineSTL processing a new point for a series with one seasonality period. A series of online filters are employed to extract the trend and seasonality. trend3, seasonality2, and residual are the final trend, seasonal, and residual.}
\label{fig:onlineSTL_on_sampledata}
\end{figure*}

\section{Filters}
\label{filter}
OnlineSTL applies a sequence of online filters iteratively. We describe two important filters used extensively throughout.

\subsection{Trend Filter}
\label{trend-filter}

Trend filters capture implicit, slowly varying patterns in the trend of a scalar time series $X = \{X_1, X_2, \cdot \cdot \cdot, X_k\}$. They have applications in finance, application monitoring systems \cite{asap}, meteorology, and medical sciences. See \cite{l1trend} for more applications. There are many trend filtering techniques. Some of these include moving averages \cite{movingavg} and exponential smoothing \cite{expavg}. We focus on a variant of symmetric kernel smoothing, a technique which was employed in original STL \cite{stl}. A symmetric kernel filter looks at a neighborhood of a point (in both past and future) in order to produce a weighted estimate for the current point. Naturally, this means that symmetric smoothing is a batch method. Local regression \cite{stl}, spline fitting \cite{spline},  Hodrick-Prescott(H-P) filtering \cite{hp1},  $\ell_1$-trend filtering \cite{l1trend} all fall under the category of batch algorithms.

A variety of kernel filters can be used to filter trend in OnlineSTL and the choice is rather flexible. Some basic filters include running mean \cite{movingavg}, running median \cite{arce2005nonlinear}, nearest neighbors \cite{friedman2001elements}, and tri-cube \cite{stl} to parabolic filters like Epanechnikov quadratic kernel \cite{epanechnikov1969}. These filters have their advantages and disadvantages. For ex., nearest neighbors is simple to compute but produces a bumpy trend, while the Epanechnikov and tri-cube filter produce smooth trends \cite{friedman2001elements}. The Epanechnikov filter is non-differentiable on the limit points of its support, while the tri-cube filter is doubly differentiable even on the limit points \cite{friedman2001elements}. This means that the tri-cube filter produces a smoother trend. On many datasets, smoother trends are more desirable because they are not as adversely affected by sudden shifts in the data due to anomalous values or irregular noise. For our experiments below, we will use the tri-cube filter, as 1) trend is generally assumed to be smooth and slowly changing, and 2) the tri-cube can be computed quickly via fast matrix operations. Smoothness (or roughness) is well-studied in kernel-smoothing literature and it is typically controlled by the window size \cite{Hansen}. Certain datasets may exhibit non-smooth trend behavior and therefore the tri-cube filter may not be a good fit. We provide more details along with an adversarial example in Section 7.3.2.

\subsubsection{Tri-cube Kernel filter} \label{sec:tri-cube}
From here on, when we refer to a trend filter, we mean the tri-cube kernel filter. We first attach a weight to each data point of the time series. The weight is computed in a neighborhood of some incoming point $X_t$ and is a function of data points in the neighborhood. The size of the neighborhood is defined as the \textit{window}. In the online setting, we look at points which are at most $\lambda$ time steps in the past. For example, for the incoming point $X_t$, we consider all points in the past $\{ X_{t- \lambda +1},\cdot \cdot \cdot, X_t \}$. In a batch setting, we may look at the points on both sides of $X_t$, i.e., $\{X_{t-\frac{ \lambda }{2}} \cdot \cdot \cdot X_{t-2}, X_{t-1}, X_{t}, X_{t+1}, X_{t+2} \cdot \cdot \cdot X_{t+\frac{ \lambda }{2}}\}$. Applying the tri-cube kernel filter on the former sequence is called \textit{non-symmetric} smoothing, while applying the same filter on the latter series is referred to as \textit{symmetric} smoothing. Furthermore, points closer to $X_t$ are given more weight than points farther away. The trend of $X_t$ is computed as the weighted average of all points in the neighborhood. Rigorously, suppose $X_t$ is some incoming/most recent point. Let $\lambda \in \mathbb{N}$ be the given window. The tri-cube kernel filter is a function $W: \mathbb{R} \rightarrow (0, 1]$ defined as:
\begin{equation}
W(u)=
\begin{cases}
  (1-({u}^3))^3, & \text{if }
       \!\begin{aligned}[t]
       0 \le u <1
       \end{aligned}
\\
  0, & \text{otherwise}
\end{cases}
\end{equation}
The doubly differentiable nature of the function allows us to compute smooth trends \cite{friedman2001elements}. Now given $W(u)$ and the most recent point $X_t$, we compute a weight for each element $X_i$, $t-\lambda \le i \le t$ w.r.t $X_t$. The weight is computed by using the timestamp of the point $i$ and $t$. The neighborhood weight for a point $X_i$ with respect to $X_t$ can then be computed as  $W(\frac{|i - t|}{ \lambda })$. For fast computation, we pre-store these neighborhood weights (kernels) for a given window $\lambda$ and the sum of each kernel. Namely, we store a vector $k_\lambda := \{w_k\}^{\lambda}_{k=1}$, where $w_k := W(\frac{|\lambda - k|}{\lambda})$. $k_\lambda$ is constant throughout the entirety of the algorithm and is completely determined by $\lambda$.
When a new point $X_t$ arrives with timestamp $t$ , we simply compute the dot product of the pre-stored kernel with the $\lambda$ previous points $X_{t - \lambda + 1 : t} := \{X_k\}_{k=t - \lambda + 1}^{t}$. For example, in computing the initial estimate of the trend , say $y_t$ of point $X_t$, we apply the non-symmetric trend filter $TF(\cdot, \cdot)$ on $X_{t}$, using the previous $\lambda$ points. Formally, when we say we apply the \textit{non-symmetric trend filter} $TF(\cdot, \cdot)$  on the latest point $X_t$, we mean:

$$y_t = TF(k_{\lambda}, X_t) = \frac{\langle k_\lambda \cdot X_{t - \lambda + 1 : t} \rangle}{||k_\lambda||_1}$$

Unlike symmetric kernel filters, non-symmetric  filters lags the actual trend. This is because
knowledge of future points (in case of symmetric filters) corrects the weight of a point.
To mitigate this problem, we extract trend iteratively, e.g., extracting trend $TF(k_\lambda, X_t)$ on data point $X_t$, then extracting remaining trend again via $ TF(k_\lambda, X_t - TF(k_\lambda, X_t))$. The resulting net trend is  $TF(k_\lambda, X_t)$ +   $TF(k_\lambda, X_t - TF(k_\lambda, X_t))$.



\subsection{Seasonality Filter}
\label{seasonal_filter}
Seasonality filters extract seasonal components from a time series. Fourier coefficient-based seasonality filters are used to obtain seasonal coefficients in popular packages such as Prophet \cite{prophet}, TBATS \cite{tbats}, etc. However, these approaches cannot handle variation in seasonality over time as the coefficients are fixed. Another common approach is to consider the mean of detrended data at different seasonal values. For example, if seasonality is yearly for data arriving monthly, then the mean of a fixed month across different years can be a good estimate of seasonality for that month. The disadvantage of using the mean, however, is that each point has equal weight in the average. When estimating the seasonal component for the most current point in the online case, we want to place more weight on datapoints which have occurred recently rather than further into the past. Hence, OnlineSTL employs exponential smoothing for its seasonality filters. At the same time, we allow seasonal components to change over time. Depending on the data, a user may place higher weight on the latest point if seasonality is changing fast. 

For simplicity, assume we are given a detrended series, $D = \{d_1, ... d_t\}$, where $t$ is the latest timestamp, and we wish to compute the seasonal component for the latest term $d_t$. $D$ is usually obtained by applying  trend filters above. Theoretically, if we were able to remove all the trend from $X_t$, then each $d_t$ would be the  seasonal component for $X_t$ in the decomposition. However, this is almost never the case. It is likely $D$ may still contain trend along with some noise after removing the initial trend estimate using  trend filters. As an example, one can think of series $D$ as the resulting series obtained after removing an initial estimate of the trend (trend1 in sections below) from the original series $\{X_k\}_{k=1}^{t}$. 

Before we define what a seasonality filter is, we will need to define some additional terms. For the most recent time $t$, let $\mathcal{F}_t = \{C^{D}_k(t)\}_{k=1}^{m}$ be a family of sets, where $m$ is the seasonality period, and $C^{D}_k(t)$ is the \textit{k-th cyclic subseries}, $C^{D}_k(t) = \{ d_r | 1 \le r \le t, r \ \textrm{mod} \ m = k \}$. It is clear that $C^{D}_k(t)$ partitions $D$. 

For each $C^{D}_k(t)$, we can then exponentially smooth over the set $C^{D}_k(t)$ via a \textit{seasonality filter}. If $C^{D}_k(t) = \{ d_{k}, d_{k + m}, d_{k + 2m}, d_{k+3m}, ... \}$, then applying a seasonality filter on $C^{D}_k(t)$ will give us the smoothed set $\mathcal{SC}^{D}_k(t)$ defined recursively as $\mathcal{SC}^{D}_k(t) = \{c_{k + im} | c_{k + (i+1)m} = \gamma d_{k + (i+1)m} + (1 - \gamma) c_{k + im} , i \in \mathbb{Z}_+, k+im \le t , c_k = d_k\}$, where $\gamma$ is the exponential smoothing factor. For each cyclic subseries, we apply the seasonality filter, and the resulting series obtained from rearranging the elements over all cylic subseries in order is the seasonal series obtained from $D$. In other words, $\mathcal{C}^{D} = \{ c_1, c_2, c_3, .... c_t \}$ is the seasonal series obtained from $D$. 

\section{Algorithm}
At its core, OnlineSTL is a simple algorithm which applies non-symmetric trend and seasonality filters alternately on the most recent window of points. 
The algorithm consists of two parts: a single offline initialization phase and a persistent online phase. In the offline phase, we run STL algorithm \cite{stl} on an initial batch of points to initialize a set of arrays we will maintain during the online phase. In the online phase, for each new incoming point, we apply a sequential set of filters as shown in Figure \ref{fig:onlineSTL_on_sampledata}.

\subsection{Notation}
The following notation will be needed.

\noindent
\begin{enumerate}[topsep=0pt, partopsep=0pt, leftmargin=*, align=left, labelindent=0pt]
  \item $X = \{ X_1, X_2, X_3, ... \}$ := the original time series, each point arriving one after another. Each $X_i \in \mathbb{R}$ 
  \item $m_p$ , $1 \le p \le k$ := user specified seasonality periods. 
  \item $m \in \mathbb{N}$ := $\max \{m_1, m_2, ..., m_k \}$, the maximum seasonality. 
  \item $\mathcal{A} \in \mathbb{R}^{4m}$ := array to store the latest $4m$ elements of $X$. Can be thought of as a sliding window of the latest $4m$ elements.
  \item $\mathcal{K}_p \in \mathbb{R}^{4m}$ := the seasonal series of period $m_p$  obtained after removing the initial trend of the latest $4m$ points.
  \item $\mathcal{E}_{p, S} \in \mathbb{R}^{m_p}$ := running smoothed seasonality estimates for $m_p$ of the initially detrended series, $\mathcal{E}_{p, T} \in \mathbb{R}^{m_p}$ :=  running smoothed seasonality estimates for $m_p$ of the series obtained after removing initial trend and trend of seasonality. The latter is a more accurate estimation of the seasonality components since $\mathcal{E}_{p, T}$ is obtained from a detrended series which also removes trend of the seasonality $m_p$.
  \item $\gamma \in [0, 1]$ := smoothing parameter used for seasonality filtering. Set to $0.7$ in implementation. 
  \item $\mathcal{D} \in \mathbb{R}^{m}$ := array which represents the completely deseasonalized series of the last $m$ elements
  \item \Call{UpdateArray}{$X$, $y$} := a simple operation on a
  circular array that replaces the oldest element in $X$  and  with $y$. 
\end{enumerate}

\subsection{Initialization:}
We need to initialize the following arrays: \textbf{1)} $\mathcal{A}$, \textbf{2)} $\mathcal{K}_p$, \textbf{3)} $\mathcal{E}_{p, S}$ $\forall 1 \le p \le k$, \textbf{4)} $\mathcal{E}_{p, T}$ $\forall 1 \le p \le k$, and \textbf{5)} $\mathcal{D}$. These arrays will constantly be updated during subsequent update phases. First, initialize $\mathcal{A}$ as the first $4m$ points. Initialization is similar to  offline STL \cite{stl}.

\begin{itemize}[leftmargin=*]
    \item For each point in $\mathcal{A}$, apply symmetric trend filter  on the point with a window of $2m_p$ and subtract the resulting trend obtained from this point. Call the resulting initially detrended series $T_1$.
    \item For each cyclic subseries in $T_1$, apply seasonality filter to get the smoothed cyclic subseries. Save the last value of $r$'th smoothed cyclic subseries in $\mathcal{E}_{p, S}[r], 0 \le r \le m_p - 1$. 
    \item Initialize $\mathcal{K}_p$ as the union over all smoothed cyclic subseries in order of time. This is a seasonal series obtained from $T_1$.
    \item For each point in $\mathcal{K}_p$ apply a symmetric trend filter of window size $\frac{3m_p}{2}$ and subtract the trend obtained from corresponding element in $T_1$. Define the series obtained as $D_5$. This series is the series obtained after removing initial trend and trend of the seasonal series from above. 
    \item Same exact procedure as 3 steps above. For each cyclic subseries in $D_5$, apply seasonality filter to get the smoothed cyclic subseries. Save the last value of $r$'th smoothed cyclic subseries in $\mathcal{E}_{p, T}[r], 0 \le r \le m_p - 1$. During update phase, we will be outputting our $S_{p, i}$ values from $\mathcal{E}_{p, T}[r]$.
    \item Let $Q$ be the union over all smoothed cyclic subseries obtained from above step, restricted to the latest $m$ points. Update $\mathcal{D}$ as $\mathcal{D} := \mathcal{D} - Q$, where the initial value of $\mathcal{D}$ is $\mathcal{A}$. 
    \item Return to the first step and iterate to the next $m_p$. 
\end{itemize}

\subsection{Update:}
 The exact update procedure is given in Algorithm \ref{alg:update}. For every arriving point $X_i$, we compute the seasonal components $S_{p, i}$, by 1) applying a trend filter on latest points $\mathcal{A}$ (including $X_i$) using a window proportional to $m_p$ to get $t_1$, 2) detrending $b$ by $t_1$, 3) updating the latest element of the seasonal series, $\mathcal{K}_p[4m-1]$, with the $i$th/newest point of the initially detrended series via seasonality filtering, 4) detrending the initial seasonal series $\mathcal{K}_p$ to get $t_4$, and then finally computing our final seasonal component on the value $b - t_1 - t_4$, which is a value without the initial trend and trend of the initial seasonality estimate. Once all seasonality estimates have been obtained, we update $\mathcal{D}[m-1]$ using $b$, the array which keeps track of the completely deseasonalized series, and compute the final trend on this series, $T_i$. An example of this procedure is given in Figure \ref{fig:onlineSTL_on_sampledata} for a time series with just one seasonality. 

\begin{algorithm}[!ht]
  \caption{OnlineSTL, Online Update} 
  \label{alg:update}
  \begin{algorithmic}[1]
  \State \textbf{Input: }new point $X_i$, $i>4m$.
  \State \Call{UpdateArray}{$\mathcal{A}$, $X_i$} \Comment{latest $4m$ points}
  \State $b := X_i$
  \For{$1 \le p \le k$}
    \State $t_1 := TF(k_{4m_p}, \mathcal{A}[4m-1])$
    \label{update:l1}
    \Comment{non-sym \textbf{trend filter}}
    \State $d_1 := b - t_1$ 
    \label{update:l2}
    \State $r := (i-1) \ \textrm{mod} \ m_p$
    \label{update:l3}
    \State $\mathcal{E}_{p, S}[r] = \gamma \cdot d_1 + (1 - \gamma) \mathcal{E}_{p, S}[r] $
    \label{update:l4}
    \Comment{\textbf{seas. filter}, update $\mathcal{E}_{p, S}$}
    \State \Call{UpdateArray}{$\mathcal{K}_p$, $\mathcal{E}_{p, S}[r]$} 
    \label{update:l5}
    \State $t_4 := TF(k_{3m}, \mathcal{K}_p[4m-1])$ 
    \label{update:l6}
    \Comment{non-sym \textbf{trend filter}}
    \State $d_5 := b - t1 - t4$ 
    \label{update:l7}
    \State $\mathcal{E}_{p, T}[r] = \gamma \cdot d_5 + (1 - \gamma) \mathcal{E}_{p, T}[r]$
    \label{update:l8}
    \Comment{\textbf{seas. filter}, update $\mathcal{E}_{T}$}
    \State $b := b - \mathcal{E}_{p, T}[r]$ \Comment{deseasonalize $b$ for next iteration}
    \label{update:l9}
  \EndFor
  \State
  \State \Call{UpdateArray}{$\mathcal{D}$, $b$} 
  \State $T_i = TF(k_{m}, \mathcal{D}[m-1])$ \Comment{non-sym \textbf{trend filter}, final trend}
  \State $S_{p, i} = \mathcal{E}_{p, T}[r]$ \Comment{final seasonality}
  \State $R_i = X_i - T_i - \sum_{p=1}^{k} S_{p, i}$ \Comment{final residual}
  \end{algorithmic}
\end{algorithm}

\noindent
 The hyperparameters we need to set in Algorithm \ref{alg:update} are the size of the past window $\mathcal{A}$, which determines the sizes of the trend filters as they should never be larger than the total number of points stored in the past window, and the seasonality smoothing parameter $\gamma$. The runtime of a single update is $O(km)$. For each seasonality, the most expensive computation in the for loop are the two trend filters, which both take $O(m)$ time. Hence, increasing the window size adversely affects the throughput but gives a more accurate estimate of the trend. On the other hand, small window sizes will cause the trend of the seasonal series to not be captured, causing a waterfall effect of incorrectly computing $d_5$, the detrended series. To avoid this issue, the window size should be at least greater than $2m$. In general, if the underlying data exhibits low(high) seasonal fluctuations, then it suffices to use a smaller(larger) window size closer(farther away from) to $2m$. Due to the iterative refinement of trend and seasonal estimates in the update procedure, we also use smaller windows for subsequent trend filters, as this further improves throughput without a noticeable penalty in our accuracy experiments. Finally, at the expense of speed, we can also add more trend filters in the for loop to compute a more accurate trend by iteratively refining the trend as explained at the end of Section \ref{sec:tri-cube}. In our accuracy experiments below, we find using two trend filters is sufficient to give competitive results.

If $\gamma \rightarrow 0$, then the effect of $d_1$ and $d_5$ on seasonal estimates is very small(lines \ref{update:l4} and \ref{update:l8}), leading to seasonal estimates which will be relatively constant and smooth. On the other hand, if $\gamma \rightarrow 1$, then seasonal estimates will be more affected by the most recent point, resulting in seasonal series with more variability. For ex., anomalous points will affect the movement of the seasonal estimates more. We found that setting the size of the past window to $4m$ and $\gamma = 0.7$ already gives competitive results in accuracy while maintaining high throughput. That being said, these values are not the only set of values which will give good results due to the trade-offs discussed above, and other values will certainly work. 

\section{Deployment of OnlineSTL}
Internally, OnlineSTL is implemented in Scala and deployed on top of Apache Flink \cite{flink}. A typical
data stream contains hundreds of thousands of time series each maintaining a 
separate key. Flink automatically balances the load among its workers 
given a user defined parallelism parameter. For example, we can ingest 20,000 time series into Flink and set parallelism to 100. Then roughly every worker (task slot
in case of Flink) will receive 200 time series. 

{\bf Throughput metric:} The throughput of an algorithm is measured as the number
of data points an algorithm can process per second. If data arrives
at 1-sec intervals, then throughput of the algorithm should be at least 
one record/sec. When we measure throughput, we set the rate of ingestion to be high 
(usually millions of data points per second per thread). This ensures 
the data source is not the throughput bottleneck. Internally, this causes
back pressure since the source produces data faster than downstream operators can consume. Other elements  can also affect throughput, such as checkpointing, file system, network buffers, etc.

For testing, we deployed Flink's instance on AWS EC2 c5.9xlarge instance with 128 CPUs and 160 GB RAM. Flink uses 4 taskmanagers with 32 task
slots each.  We set Flink's parallelism to 120 to obtain our throughput benchmarks. 
Data streams are generated using an event generator, which simulates real metrics Splunk customers may see. We generate unbounded data streams containing 100K time series metrics, which is a realistic number of metrics clients may need to monitor at the same time. For single node experiments, we generate a single time series used across all benchmark algorithms. Checkpointing is disabled to prevent throughput bottlenecking. Throughput is measured as the average over all task slots and memory numbers are reported for the whole instance in Table 
~\ref{table:performance_splunk_internal}. 

\begin{table}[h]
\centering
\caption{Performance Benchmarks on Apache Flink. Throughput is given per task slot per second.}
{\fontsize{9}{9}\selectfont
\begin{tabular}[l]{ cccccc } 
\toprule
 Seasonality & Throughput & JVM Heap & Total events/s \\
\midrule
10 & 85K & 24GB & 10.1M \\
100 & 69K & 28GB & 8.3M \\
1000 & 25K & 36GB & 3.0M \\
10000 & 3.6K & 108GB & 440K \\
\bottomrule
\end{tabular}
}
\label{table:performance_splunk_internal}
\end{table}

Table \ref{table:performance_splunk_internal} gives high throughput values for DevOps metrics monitoring. A typical time series at Splunk is often aggregated at minutely resolution or higher(for ex., real time CPU usage every $x$ seconds), meaning that weekly seasonality is $O(10,000)$. The number of metrics tracked are in order of millions. It is imperative to have high throughput in order to lower operational costs and provide
quick results for downstream operators. A throughput of
100 records/sec would need 10 times more resources than a throughput of 1000 records/sec. The expected behavior that throughput decreases as seasonality increases is due to additional compute and memory required by filters. While throughput decreases for larger seasonalities, it is clear OnlineSTL can still handle long seasonalities of period 10K with total throughput of around half a million. This means for metrics monitoring at high resolution with hundreds of thousands of metrics, OnlineSTL maintains high performance.

In a recent use case, one of our customers applied classical STL on weekly data aggregated at 30 minute intervals. STL is often used as a pre-processing operator, whereby results are then passed into anomaly detection operators. Since classical STL is a batch algorithm, our customer faced the problem of having to retrain each metric whenever a new point arrived for that metric, making scalability over a few thousand metrics costly as time and compute costs grew. OnlineSTL provided a fast streaming solution which solved both problems as high throughput lowered server time costs which in turn lowered dollar costs.


\section{Experiments}
To paint a more accurate picture of OnlineSTL's high throughput, we benchmark OnlineSTL against other algorithms in an isolated environment on a single node without throughput interfering factors from checkpointing or differing file systems.

\subsection{Compared Algorithms} \label{compared_algorithms}
We evaluate OnlineSTL against multiple techniques in batch and online-counterpart modes. We can naturally construct an online counterpart of any batch algorithm by adding the newest point in the time series and then applying the batch algorithm on some past window of points. These batch algorithms are given below:

\begin{itemize}[leftmargin=*]
\item {\bf STL}\cite{stl}: performs decomposition based on Loess smoothing.
\item {\bf TBATS}\cite{tbats}: decomposes a series into trend, level, seasonality and residual. Here, sum of trend and level is equivalent to standard trend used in OnlineSTL.
\item {\bf STR}\cite{str}: jointly produces decomposition using regression. 
\item {\bf SSA}\cite{ssa}: uses  SVD to produce decomposition.
\item {\bf RobustSTL}\cite{robuststl}: optimization technique which uses robust loss function and sparse regularization.
\item {\bf Fast RobustSTL}\cite{frobustSTL}: an extension of RobustSTL which allows for multiple seasonalities. 
\end{itemize}

\begin{figure}[h]
\centering
\includegraphics[scale=0.45]{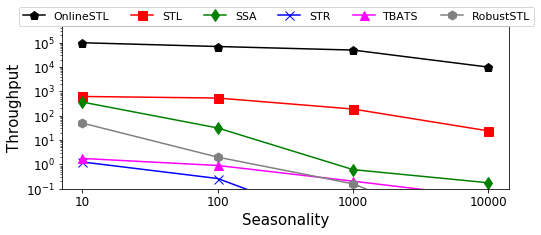}
\caption{Throughput comparison for different seasonalities: Online STL vs. online version of batch algorithms }
\label{Fig: throughput}
\end{figure}

\subsection{Single Node Throughput}
Single node throughput results are shown in Figure \ref{Fig: throughput}. We compare OnlineSTL to the online versions of batch algorithms. For fair comparison, we process data points in a sliding window of size $4m$, as this is the window OnlineSTL uses for its filters. Our single node environment is an Intel 2.4GHz Core i9 with 16 GB RAM. From the graph, OnlineSTL is at least 100 times faster than any other algorithm on high ingest datasets with large seasonalities. Note how classical STL is still the second fastest algorithm. What is more worthy of note is that OnlineSTL scales well as seasonality increases, whereas other algorithms apart from classical STL scale much worse, experiencing dramatic drops in throughput as seasonality increases. 

\subsection{Quality of Decomposition, Accuracy}
Finally, we benchmark the decomposition quality of the aforementioned algorithms(both batch and online modes) on several datasets. Measuring the quality of decomposition quantitatively is a tricky affair. Having zero residual does not imply the decomposition is good. If an algorithm returns trend as the original data set, then the residual is zero, but the decomposition would be undesirable since seasonal terms were not captured. To avoid this problem, we use smoothness of trend along with the residual to measure quality of decomposition. 

In time series analysis, true trend is assumed to be slowly changing and smooth. If the extracted trend of some decomposition is close to the original data and hence captures most of the seasonality and residual, then the extracted trend is not smooth. On the other hand, trend which does contain seasonal effects or residual is smoother as the trend will not contain any oscillations or randomness that seasonal effects or residuals may contain respectively. This means smoothness of trend and residuals will often move in opposite directions. If trend is more smooth (lower smoothness of trend), then residuals will increase, while if trend is less smooth (higher smoothness of trend), then residuals will decrease. The ideal decomposition should both be smooth and have low residual.  We use MASE of Residual and trend smoothness as two measures to quantify quality of a decomposition. 

\noindent {\bf Mean Absolute Scaled Error (MASE) of Residual: } 
 MASE is a scaled version of mean absolute error and is independent of the scale of the data. It is used to compare error across datasets and is given in Eq. \ref{MASE}  \cite{Hyndmanmase1, Hyndmanmase2}. 

\begin{equation} \label{MASE}
    mean \left( \frac{|e_j|}{ \frac{1}{T-1} \sum_{t=2}^T |X_t - X_{t-1}| } \right)
\end{equation}

Here, $T$ is the size of the time series, $e_j$ is the error of the $j$'th forecast and $X_t$ are the values in the time series. The average is taken over the number of forecasted values. Intuitively, MASE computes the forecast error of the proposed method over the average error of using $X_{t-1}$ as the forecasted value for $X_t$. When MASE is less than 1, the proposed method gives on average smaller errors than the one-step errors of the naive method \cite{hyndman2006another}. On the other hand , when MASE is large, then the proposed algorithm is worse than the naive algorithm.

\noindent {\bf Smoothness of trend:} We measure smoothness of trend as the standard deviation of the first order difference of the trend series \cite{variance}. Lower(Higher) values indicate a smoother(rougher) series. 

\subsubsection{Real Datasets:}
 We tested each algorithm on a total of 5 real datasets (described in  Table \ref{decomp-experiments}). These datasets come from a variety of sources and we give a short description of each below:
\begin{itemize}[leftmargin=*]
    \item \textbf{Bike sharing \cite{bike_sharing, UCI}:} Bike sharing dataset from UCI repository. Dataset contains daily bike sharing rental totals from Jan. 2011 to end of Oct. 2012. 
    \item \textbf{Daily female births \cite{tsdl}:} Dataset from datamarket and time series data library (tsdl). Dataset contains daily number of female births from Jan. 1959 to Dec 1959. 
    \item \textbf{Elecequip \cite{fpp2}:} Monthly manufacturing of electrical equipment dataset from R fpp2 package. Dataset contains monthly data from Jan. 1996 to Mar. 2012. 
    \item \textbf{Min temperature \cite{tsdl}:} Daily minimum temperature dataset from tsdl. Dataset contains daily data starting from 1981. 
    \item \textbf{Internet traffic \cite{tsdl}:} Hourly internet traffic dataset from datamarket and tsdl. It contains hourly internet traffic from 11 Eurpean cities. 
\end{itemize}

\noindent {\bf OnlineSTL vs batch results:} Table \ref{decomp-experiments} contains results for each  algorithm.
We compare OnlineSTL with different batch algorithms. In the top section, we observe that the MASE of OnlineSTL is very competitive compared to the batch algorithms. In fact, it 
performs extremely well across most datasets, and always places in the upper half of algorithms with the lowest MASE (with the exception of the min temperature dataset, in which it places 6th out of 11). In comparing trend smoothness, OnlineSTL is also extremely competitive and is always in the upper half of algorithms with the smoothest trend, which is surprising given that it is an online algorithm. 


\begin{table*}[t]
\centering
\caption{Accuracy results over real datasets. Results are in the form of x/y, where x is the result for batch version of the algorithm and y is  for the online version of the algorithm. Results in bold (bold italic) represent the lowest MASE/trend smoothness across all batch (online) algorithms.}
\resizebox{\textwidth}{!}{
\begin{tabular}{| c |c|c| c c c c c c c | c c c c c| c |}
    \hline
    \textbf{Dataset, MASE of res.} & Size & Seasonality & \textbf{STL} & \textbf{SSA} & \textbf{STR} & \textbf{TBATS} & \textbf{(Fast) RobustSTL} & \textbf{OnlineSTL} & \textbf{OnlineSTL Rank}\\
    \hline
    Bike sharing & 730 & 7 & 0.513/0.475 & {\textbf{0.303}}/\textbf{\textit{0.286}} & 0.654/0.611 & 0.672/0.671 & 0.596/0.674 & 0.430 & 2/2\\
    
    Daily female births & 364 & 7 & 0.566/0.504 & 0.405/0.350 & 0.630/0.516 & 0.744/0.725 & {\textbf{0.322}}/\textit{\textbf{0.334}} & 0.462 &3/3 \\
    
    Elecequip & 190 & 12 & 0.243/0.271 & 0.419/0.455 & {\textbf{0.209}}/\textit{\textbf{0.252}} & 0.304/0.313 & 0.383/0.419 & 0.292 & 3/3 \\
    
    Min temperature & 500 & 7, 28 & 0.561/0.574 & 0.359/0.405 & 0.608/0.574 & 0.629/0.625 & {\textbf{0.149}}/\textit{\textbf{0.158}} & 0.396 & 3/2 \\
    
    Internet traffic & 1231 & 24, 168 & 0.857/1.074 & 0.765/0.622 & {\textbf{0.313}}/\textit{\textbf{0.236}} & 0.369/0.405 & 0.845/0.821 & 0.618 & 3/3 \\
    \hline
    \textbf{Trend Smoothness (log scale)} & & & & & & &\\
    \hline
    Bike sharing & 730 & 7 & 4.831/6.000 & 5.677/6.292 & {\textbf{3.598}}/\textit{\textbf{4.935}} & 6.982/6.988 & 5.002/5.445 & 5.378 & 4/2 \\
    
    Daily female births & 364 & 7 & -0.334/1.035 & 0.837/1.579 & {\textbf{-1.256}}/0.730 & 2.186/2.162 & -0.0177/0.709 & \textit{\textbf{0.353}} & 4/1 \\
    
    Elecequip & 190 & 12 & -0.132/0.415 & 0.217/1.020 & 0.168/0.789 & 1.272/1.410 & 0.429/0.535 & {\textbf{-0.175}} & 1/1 \\
    
    Min temperature & 500 & 7, 28 & -2.550/-1.079 & -2.708/-1.918 & {\textbf{-2.793}}/-0.902 & 1.031/1.032 & -1.561/-1.879 & \textit{\textbf{-2.18}} & 4/1 \\
    
    Internet traffic & 1231 & 24, 168 & 16.305/19.659 & 16.669/18.872 & {\textbf{14.860}}/20.745 & 22.682/23.580 & 18.807/20.756 & \textit{\textbf{18.245}} & 4/1 \\
    
    \hline
    \textbf{Algorithm Runtime (seconds)} & & & & & & &\\
    \hline
    Bike sharing & 730 & 7 & 0.0038/1.491 & 0.0108/6.143 & 1.803/829.571 & 1.875/349.017 & 4.443/12.812 & \textit{\textbf{0.0033}} & 1/1 \\
    Daily female births & 364 & 7 & 0.0031/0.819 & 0.0089/3.706 & 1.227/376.081 & 0.836/172.345 & 0.650/5.979 & \textit{\textbf{0.0016}} & 1/1 \\
    Elecequip & 190 & 12 & 0.0034/0.493 & 0.0072/1.4363 & 1.351/192.448 & 1.517/113.423 & 0.220/5.510 & \textit{\textbf{0.0017}} & 1/1 \\
    Min temperature & 500 & 7, 28 & 0.0091/2.551 & 0.0206/6.947 & 3.045/1122.953 & 2.702/657.664 & 2.464/175.649 & \textit{\textbf{0.0053}} & 1/1\\
    Internet traffic & 1231 & 24, 168 & 0.0134/6.363 & 0.112/32.390 & 18.342/7995.889 & 7.179/2486.139 & 23.410/3240.716 & \textit{\textbf{0.0112}} & 1/1\\
    
    \hline
\end{tabular}
}

\label{decomp-experiments}
\end{table*}

\noindent {\bf OnlineSTL vs batch-online results:} In these experiments, we
compare the online counterpart of each batch algorithm to OnlineSTL. Suppose we have an unbounded data stream, and the most recent time is $t$. An online-batch algorithm will consider all points from $X_1, X_2, \cdot \cdot \cdot, X_t$, compute a decomposition on these $t$ points, and then output the decomposition of $X_t$. When $X_{t+1}$ arrives, it will compute the decomposition on $X_1, X_2, \cdot \cdot \cdot, X_{t+1}$, and then output the decomposition of $X_{t+1}$. This continues until the entire dataset has been seen. We use a sliding window of $4m$ past points for computing online decomposition for these batch algorithms.


Comparing the MASE of OnlineSTL with the online counterparts of the batch algorithms, OnlineSTL remains extremely competitive as shown in the top section of Table \ref{decomp-experiments}.  OnlineSTL ranks either second or third in either batch or online categories, and it ranks either 3/11, 4/11 or 5/11 overall among all algorithms in MASE for a fixed dataset. If we compare the MASE scores of batch algorithms with their online counterparts, we observe that some of them improve, which is a bit counter intuitive. The intuition is that computing decomposition on limited data should produce worse results. This can be explained due to overfitting: recall that if a decomposition returns the trend as the original data point, then the residual will be zero despite the decomposition being worthless. This behavior of overfitting is captured in the middle section of Table \ref{decomp-experiments}  where we compare trend smoothness. We observe trend smoothness over all batch algorithms has worsened. Recall that a challenge for any online algorithm is to produce smooth trends. OnlineSTL has the best trend smoothness among all the online algorithms, producing the smoothest trend in four out of five datasets and the second smoothest in the fifth dataset. Finally, we mention that it is expected that the trend smoothness of OnlineSTL will compare unfavorably to batch methods, as these methods can look to points in the future to smooth out the current point.

\begin{figure}[t]
  \centering
{\includegraphics[scale=0.27]{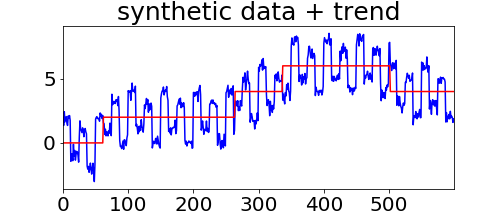}\label{fig:synth_trend}}
  \caption{Synthetic data with five trend changepoints}
  \label{Fig: synthetic}
\end{figure}

\subsubsection{Synthetic Dataset:}
Previously, we observed that OnlineSTL performed quite well on real datasets. Here we generate an adversarial synthetic dataset and benchmark the above algorithms against the true trend and seasonality. The generated dataset contains 750 points, with seasonal components of length 25 and 50. Note that the true trend of the synthetic dataset has rather abrupt changes at 5 random points.  The final series is constructed by adding the trend and seasonal components together with random Gaussian noise. The generated time series is shown in Figure \ref{Fig: synthetic}.

In Table \ref{table:synthetic_MASE}, we see that OnlineSTL captures MASE of both seasonalities well, and is competitive and in the middle of pack. With respect to trend, OnlineSTL has the smoothest trend, but also a relatively high MASE of trend. This is because true trend is discontinuous and non-smooth, which causes the tri-cube kernel to not catch up quickly due to these sudden shifts. This behavior in the tri-cube is precisely why trend is assumed to be smooth and slowly changing, as this allows the tri-cube kernel to quickly adjust to changes in the trend. A different choice of kernel filter may adapt better to these abrupt changes. The trade-off between MASE and smoothness is also highlighted here, as we see that trend smoothness is low while MASE of trend is high. The choice of kernel filter ultimately depends on the underlying data, and prior knowledge helps in identifying correct filters to choose. In this example, we showed parameters need to be chosen properly in order to respect the underlying structure in the data.

\begin{table}[h]
\centering
\caption{Comparison of different algorithms on an adversarial synthetic  series in Figure \ref{Fig: synthetic}.}
{\fontsize{9}{9}\selectfont
\resizebox{\columnwidth}{!}{
\begin{tabular}[l]{ cccccccc } 
\toprule
 & Algorithm & MASE, s=25  & MASE, s=50 & MASE trend & Trend smoothness \\
\midrule
&  STL & 0.080 & \textbf{0.078} & 0.168 & 0.020\\
& SSA & 0.971 & 0.899 & 0.193 & \textbf{0.018}\\
Batch & STR & 0.063 & 0.115 & 0.211 & 0.019\\
& TBATS & \textbf{0.062} & 0.127 & 0.220 & 0.483\\
& Fast RobustSTL & 1.095 & 0.366 & \textbf{0.074} & 0.114\\
\midrule
&  STL & 0.083 & \textit{\textbf{0.091}} & 0.210 & 0.046\\
& SSA & 0.992 & 0.956 & 0.348 & 0.035\\
Online & STR & 0.134 & 0.176 & 0.259 & 0.375\\
& TBATS & \textit{\textbf{0.08}} & 0.136 & 0.225 & 0.482\\
& Fast RobustSTL & 1.088 & 0.369 & \textit{\textbf{0.193}} & 0.075\\
\midrule
& OnlineSTL & 0.279 & 0.236 & 0.564 & \textit{\textbf{0.018}}\\
\bottomrule
\end{tabular}
}}
\label{table:synthetic_MASE}
\end{table}

\section{Conclusion and Future work}
Time series decomposition is a well studied problem in the batch setting, where recent research explored detection of anomalies, prompt response to seasonality and trend shifts as well as improved computational efficiency. However, scalability of these algorithms is rarely addressed. Conventional batch algorithms are too slow and cannot be used for real-time DevOps metrics monitoring. OnlineSTL, a novel algorithm which decomposes time series online, incrementally updates parameters and maintains decomposition accuracy for metrics monitoring use cases while providing significant improvement in throughput in time-series exhibiting large seasonal patterns. Internally, OnlineSTL has been successfully deployed for over a year and performs well on long seasonality. At the same time, it can handle high resolution data (ingestion rate is in sec) along with ability to support multiple seasonality such as daily and weekly. In the future, we  plan on investigating streaming techniques for outlier resilience, seasonality shifts, non-integral seasonality and changing seasonality. 



\newpage
\balance
\bibliographystyle{ACM-Reference-Format}
\bibliography{main}

\end{document}